\documentclass[english,a4paper,12pt,twoside]{article}

\usepackage[T1]{fontenc}
\usepackage[utf8]{inputenc}
\usepackage[english]{babel}
\usepackage{microtype}
\usepackage{textcomp}

\usepackage[letterpaper,top=2cm,bottom=2cm,left=3cm,right=3cm,marginparwidth=1.75cm]{geometry}
\usepackage{setspace}
\usepackage{amsmath,amssymb,amsthm}
\usepackage{graphicx}
\graphicspath{{Figures/}{Images/}{../Images/}}
\usepackage{subcaption}  % for subfigure (a), (b), ...
\usepackage[labelfont=bf]{caption}
\usepackage{float}
\usepackage[rightcaption]{sidecap}
\usepackage{wrapfig}
\usepackage{array}
\usepackage{booktabs}
\usepackage{makecell}
\usepackage{hhline}
\usepackage{threeparttable}
\usepackage{multirow}
\usepackage[table,xcdraw]{xcolor}
\usepackage{diagbox}

\usepackage{titlesec}
\titleformat*{\section}{\LARGE\bfseries}
\titleformat*{\subsection}{\Large\bfseries}
\titleformat*{\subsubsection}{\large\bfseries}
\titleformat{\paragraph}{\normalfont\normalsize\bfseries}{\theparagraph}{1em}{}
\titlespacing*{\paragraph}{0pt}{3.25ex plus 1ex minus .2ex}{1.5ex plus .2ex}

\usepackage{enumitem}
\usepackage{adjustbox}
\usepackage{chngcntr} % for correct table/figure numbering when needed
\usepackage{xparse,nameref}
\usepackage[bottom]{footmisc} % footnotes anchored at bottom
\usepackage{subfiles}
\usepackage{fancyhdr}
\usepackage{todonotes} % comment out for final
\usepackage{natbib}
\setcitestyle{numbers}
\setcitestyle{square}
\usepackage[colorlinks=true,allcolors=blue]{hyperref}
\begin{document}

% If you use subfiles, include front page etc.

\restoregeometry

\pagenumbering{gobble}
\thispagestyle{plain}
\pagenumbering{arabic}

% ... your content ...

% \biblio % or \bibliography{References} with your chosen style

\def\biblio{} % resets the biblio command, if not here a new reference list will be produced after every chapter

\include{NHH-Frontpage}
\restoregeometry % restores the margins after frontpage
%\nocite{*} % uncomment if you want all sources to be printed in the reference list, including the ones which are not cited in the text 

\pagenumbering{gobble} % suppress page numbering
\thispagestyle{plain} % suppress header
%\clearpage\mbox{}\clearpage % add blank page

\pagenumbering{arabic} % starting roman page numbering

\title{Orientation-Free Neural Network-Based Bias Estimation for Low-Cost Stationary Accelerometers}
\author{Michal Levin and Itzik Klein\\
The Hatter Department of Marine Technologies,\\
Charney School of Marine Sciences, University of Haifa,\\
Haifa, Israel}
\date{}
\maketitle

% ------------ Introduction ---------------
% \newpage
\section*{Abstract}

Low-cost micro-electromechanical accelerometers are widely used in navigation, robotics, and consumer devices for motion sensing and position estimation. However, their performance is often degraded by bias errors. To eliminate deterministic bias terms a calibration procedure is applied under stationary conditions. It requires accelerometer leveling or complex orientation-dependent calibration procedures. To overcome those requirements, in this paper we present a model-free learning-based calibration method that estimates accelerometer bias under stationary conditions, without requiring knowledge of the sensor orientation and without the need to rotate the sensors. The proposed approach provides a fast, practical, and scalable solution suitable for rapid field deployment. Experimental validation on a 13.39-hour dataset collected from six accelerometers shows that the proposed method consistently achieves error levels more than 52\% lower than traditional techniques. On a broader scale, this work contributes to the advancement of accurate calibration methods in orientation-free scenarios. As a consequence, it improves the reliability of low-cost inertial sensors in diverse scientific and industrial applications and eliminates the need for leveled calibration.

\section{Introduction}\label{intro_sec}

Micro-electro-mechanical system (MEMS) accelerometers are widely used in navigation, robotics, motion tracking, virtual reality, biomedical, and healthcare applications due to their small size, low power consumption, and affordability \cite{cui2018calibration, ahmad2013reviews, harindranath2023systematic}. These sensors, typically configured as triaxial devices, measure specific force by detecting the movement of an internal mass within a microscopic mechanical structure \cite{titterton2004strapdown, faisal2019review}. 

MEMS accelerometers are fabricated using micro-machining techniques on silicon wafers, a process that enables their compact size, low cost, and low power consumption \cite{mi13030343, rasch2004fiber}. This miniaturization drastically reduces cost and power consumption but increases fabrication imperfections, temperature sensitivity, and stochastic noise \cite{harindranath2023systematic}. As a result, low-cost accelerometers exhibit higher errors compared to their high-grade counterparts \cite{titterton2004strapdown, lopes}.
Within the MEMS sensor category, performance can vary depending on design precision and manufacturing quality. Industrial and tactical-grade MEMS sensors often include built-in calibration and stability enhancements, while low-cost units prioritize miniaturization and affordability at the expense of long-term accuracy.

If not properly corrected, small measurement errors, especially in low-cost sensors, accumulate during this integration process, leading to significant drift in motion tracking and the estimated navigation solution. Loosely speaking, in the later, the accelerometer output is integrated once to derive velocity and twice to derive position. Consequently, any additive error in the signal is also integrated, producing a linearly growing error in velocity and a quadratically growing position drift \cite{el2008analysis, shin2002accuracy}. To remove these deterministic error components, a calibration procedure is performed prior to mission start. This step is essential to ensure reliable performance, particularly for low-cost MEMS-based accelerometers. In most applications, the bias is the most dominant and persistent source of error and must be accurately estimated for reliable navigation \cite{groves2013principles}. 

Existing calibration approaches suffer from significant practical limitations. Many require external calibration equipment, knowledge of sensor orientation, or extended multi-position recording sessions \cite{zhong2018calibration, wang2020field, cai2013accelerometer}. These methods depend on predefined models with numerous parameters and are not easily scalable or adaptable to various real-world conditions \cite{draganova2015gradient}. Furthermore, they lack robustness to hardware variability and cannot operate under minimal constraints, which is often required in field deployments \cite{belkhouche2022robust, hoang2020robust}.

Recent advances in learning-based inertial sensor and fusion have demonstrated that deep neural networks can capture complex, nonlinear error patterns directly from raw sensor data without relying on predefined physical models \cite{cohen2024inertial, chen2024deep}. In particular, these methods have demonstrated superior performance in gyroscope stationary calibration \cite{engelsman2022learning, stolero2024rapid}. Inspired by these recent advances and to overcome the practical limitations of existing accelerometer calibration approaches, we offer a learning-based method that replaces analytical modelling in regressing the accelerometer bias in practical real-world scenarios where the orientation is unknown. To this end, we propose OFBENet, an end-to-end orientation-free neural network designed to regress accelerometer biases directly from raw stationary data. 
The key contributions of this paper are:
\begin{enumerate}
    \item {A model-free accelerometer calibration approach for practical real-world scenarios with unknown sensor orientation and without the need for rotating the sensor to different positions.}

    \item{Development of OFBENet, an end-to-end orientation-free neural network for accelerometer bias estimation. OFBENet includes a simple yet efficient neural network that learns directly from raw time-series data, capturing nonlinear error patterns without prior knowledge of sensor orientation.}

    \item{A \href{https://github.com/ansfl/OFBENet}{GitHub repository} containing our dataset and the code for implementing the OFBENet architecture. The repository serves as a comprehensive resource for researchers and practitioners interested in replicating or extending the proposed approach and to foster research in the field.}
\end{enumerate}

To evaluate the proposed approach data from multiple accelerometers, placed in different orientation, were was collected and totalling 13.39 hours of recordings under varied conditions. We demonstrate that our proposed approach outperforms two leading model-based calibration methods with higher accuracy and stable performance across devices. By eliminating the need for external equipment, predefined orientations, or long recordings, OFBENet provides a lightweight and scalable alternative to traditional calibration workflows. The method enables fast and accurate bias estimation using only stationary signals. This makes it suitable for a wide range of real-world applications, including robotics, navigation systems, and field deployments where rapid initialization is essential.

The rest of this paper is organized as follows: Section II describes the model-based
calibration approachs. Section III presents the proposed approach and Section IV provides
our analysis and results. Finally, Section V concludes this study.

% ------------ Scientific Background ---------------
\section{Background}\label{background_sec}

\subsection{Accelerometer Error Model}

Accelerometers are subject to various error sources that affect measurement accuracy. A comprehensive accelerometer error model is typically expressed as \cite{avrutov2017expanded}:

\begin{equation}
\label{acc_model_long}
\begin{bmatrix}
\tilde{f}_x \\
\tilde{f}_y \\
\tilde{f}_z
\end{bmatrix} = 
\begin{bmatrix}
b_x \\
b_y \\
b_z
\end{bmatrix} + 
\begin{bmatrix}
k_{11} & k_{12} & k_{13} \\
k_{21} & k_{22} & k_{23} \\
k_{31} & k_{32} & k_{33}
\end{bmatrix} \cdot
\begin{bmatrix}
f_{true,x} \\
f_{true,y} \\
f_{true,z}
\end{bmatrix} + 
\begin{bmatrix}
n_x \\
n_y \\
n_z
\end{bmatrix}
\end{equation}

where $\tilde{f}_i$ denotes the accelerometer output component along axis $i$, $f_{\text{true},i}$ is the true specific force acting on the sensor, $b_i$ represents the bias component, and $n_i$ is the zero-mean measurement noise, with $i \in \{x, y, z\}$ corresponding to the three orthogonal axes of the accelerometer. The elements $k_{ij}$ form the 3×3 scale and misalignment matrix, where diagonal terms represent scale factor errors and off-diagonal terms account for axis misalignment.

Among these error sources in stationary conditions, bias is typically the most dominant component and is present as a constant offset in all accelerometers during a single operation period, though it varies between power cycles \cite{titterton2004strapdown, groves2013principles}. 

\subsection{Stationary Readings}
\label{stationary_readings}
Accelerometer measurements are expressed in the sensor’s body coordinate frame, whereas the gravitational acceleration vector is defined in the navigation frame. To relate the two, a transformation matrix is employed. Using Euler angles, it is defined as \cite{groves2013principles}:

\begin{equation}
\mathbf{T}^n_b =
\begin{aligned}
\begin{bmatrix}
c_\theta c_\psi & c_\theta s_\psi & -s_\theta \\
-c_\phi s_\psi + s_\phi s_\theta c_\psi & c_\phi c_\psi + s_\phi s_\theta s_\psi & s_\phi c_\theta \\
s_\phi s_\psi + c_\phi s_\theta c_\psi & -s_\phi c_\psi + c_\phi s_\theta s_\psi & c_\phi c_\theta
\end{bmatrix}
\end{aligned}
\end{equation}

where $T^n_b$ is the transformation matrix from the navigation frame (n) to the body frame (b), and $\phi$, $\theta$, and $\psi$ represent the roll, pitch, and yaw Euler angles, respectively. 

These rotations are applied sequentially: first, the yaw (\(\psi\)) rotation around the \(z\)-axis, then the pitch (\(\theta\)) rotation around the \(y\)-axis, and finally the roll (\(\phi\)) rotation around the \(x\)-axis.

Under stationary conditions, where only gravity acts on the accelerometer, the expected accelerometer output is:
\begin{equation}\label{acc_mod_misalign}
\begin{aligned}
\begin{bmatrix}
\tilde{f} _{x} \\
\tilde{f} _{y}\\
\tilde{f} _{z}
\end{bmatrix} &= \mathbf{T}^n_b \begin{bmatrix}
0 \\
0 \\
-g
\end{bmatrix} \\
\end{aligned}
\end{equation}

where $\tilde{f}_i$ are the measured accelerometer outputs along each body axis $i \in \{x,y,z\}$, and $g$ is the gravitational acceleration magnitude under stationary conditions.

Expanding (\ref{acc_mod_misalign}) and assuming bias and white noise only, the accelerometer readings are:

\begin{equation}
\label{acc model_angle}
\begin{aligned}
\begin{bmatrix}
\tilde{f} _{x} \\
\tilde{f} _{y}\\
\tilde{f} _{z}
\end{bmatrix} = \begin{bmatrix}
\sin(\theta) \cdot g + b_x \\
-\sin(\phi) \cos(\theta) \cdot g + b_y \\
-\cos(\phi) \cos(\theta) \cdot g + b_z
\end{bmatrix}
+ 
\begin{bmatrix}
n_x \\
n_y \\
n_z
\end{bmatrix}
\end{aligned}
\end{equation}

where $\tilde{f}_i$ are the measured accelerometer outputs, $\phi$ and $\theta$ are the roll and pitch angles, $g$ is the gravitational acceleration, and $b_i$ are the bias components along each axis $i \in \{x,y,z\}$.
When the accelerometers are aligned with the local gravity vector (i.e. roll and pitch are zero), a simple average on the readings can be made to estimate the deterministic bias. Yet, in practical real-world scenarios the orientation is unknown and a simple average cannot be applied as the gravity influence cannot be determined.

\subsection{Baseline Approaches}
We review two different model-based methods capable of estimating the accelerometer bias in nonzero roll and pitch angles.
\subsubsection{Least Squares Calibration}
\label{section LS}
Least squares (LS) estimation is one of the most widely used techniques in accelerometer calibration due to its generality and flexibility. It frames calibration as a cost function minimization problem and is particularly appealing for model-free parameter estimation, as it typically does not require explicit probabilistic assumptions about the data \cite{harindranath2023systematic}. LS methods are especially effective in multi-position static calibration, where the goal is to minimize the squared error between the norm of the accelerometer measurements after bias removal and the known gravitational magnitude, based on the principle that a stationary accelerometer should output a vector with magnitude equal to $g$ \cite{FrosioI.2006AoMA}. These methods can be adapted into various forms, including weighted or recursive LS.

The LS problem can be written as:

\begin{equation}
\mathcal{L}(\mathbf{b}) = \sum_{i=1}^{N} \left( r_i(\mathbf{b}) \right)^2
\label{eq:least_squares}
\end{equation}

\noindent
\text{with } \( r_i(\mathbf{b}) \) being the residual function based on the simplified error model:
\begin{equation}
r_i(\mathbf{b}) = \left\| \tilde{\mathbf{f}}_{\text{true}, i} - \mathbf{b} \right\| - g
\label{eq:residual}
\end{equation}

\noindent
where \( \tilde{\mathbf{f}}_{\text{true}, i} \) is the accelerometer output at sample \( i \), \( \mathbf{b} \) is the bias vector, and \( g \) is the gravitational magnitude.

Since the residual function contains a squared norm involving $\mathbf{b}$, it is a nonlinear function of the bias, making the overall optimization problem nonlinear. As a result, it cannot be solved analytically using standard linear LS methods. To solve this problem numerically, the Trust region reflective (TRF) algorithm can be used. TRF is specifically designed for bounded, nonlinear least squares problems, and it is well-suited for situations in which residuals are nonlinear in the parameters. At each iteration, the algorithm constructs a local quadratic approximation of the loss function around the current estimate and solves a constrained optimization subproblem. Unlike line-search methods that fix the direction and then select a step size, TRF fixes the trust region (step size limit) first, and then determines the optimal direction within that region. This mechanism improves numerical stability and robustness, especially when far from the true minimum \cite{aggarwal2020linear}. 

The algorithm iteratively adjusts the bias estimate starting from an initial guess (zero bias) and converges to the optimal $\mathbf{b}$:

\begin{equation}
    \mathbf{b}_{i+1} = \mathbf{b}_i + \delta \mathbf{b}
\end{equation}

where  \( \delta \mathbf{b} \) is the update step. This can be written as a second-order Taylor series expansion of the cost function around \( \mathbf{b}_i \) \cite{gsl_nls, hofer2022jaxfit}:

\begin{equation}
    \mathcal{L}(\mathbf{b}_i + \delta \mathbf{b}) \approx \mathcal{L}(\mathbf{b}_i) + \nabla \mathcal{L}(\mathbf{b}_i)^T \delta \mathbf{b} + \frac{1}{2} \delta \mathbf{b}^T \nabla^2 \mathcal{L}(\mathbf{b}_i) \delta \mathbf{b}
\label{taylor}
\end{equation}

The optimal step \( \delta \mathbf{b} \) is found by minimizing Equation (\ref{taylor}) subject to the constraint \( \lVert \delta \mathbf{b} \rVert \leq \Delta \),
where \( \Delta \) is the trust region radius that limits the step size to maintain model validity.

While LS provides a mathematically grounded framework, it has inherent limitations. The optimization convergence may be sensitive to the initial guess, particularly in the presence of measurement noise or signal inconsistencies. Moreover, LS relies on multi-position datasets to provide sufficient diversity in the measurements for accurate parameter estimation \cite{harindranath2023systematic}.

\subsubsection{Mathematical Model-Based Calibration}
\label{section math model}
In addition to LS, a mathematical calibration method was suggested in \cite{won2009triaxial}. Like LS, this method assumes that for a stationary accelerometer, the norm of the three-axis acceleration measurements should equal the gravitational constant. However, instead of minimizing axis-wise residuals, this approach directly enforces this physical constraint by iteratively aligning the entire corrected signal with the expected gravitational magnitude.
At each iteration, a bias correction vector, referred to as ${Cal}_{k}$, is computed by solving a linear system of the form:

\begin{equation}
\text{Error}_{k-1} = \tilde{f}_{\text{true}, k-1} \cdot \text{Cal}_k
\end{equation}

\noindent
where
\begin{equation}
\tilde{f}_{\text{true}, k-1} = \tilde{f}_k - b_{k-1}
\end{equation}

\begin{equation}
\text{Error}_{k-1} = 
2\, b_{x,k} \, \tilde{\mathbf{f}}_{\text{true}, x, k-1} 
+ 2\, b_{y,k} \, \tilde{\mathbf{f}}_{\text{true}, y, k-1} 
+ 2\, b_{z,k} \, \tilde{\mathbf{f}}_{\text{true}, z, k-1},
\end{equation}

$\tilde{f}_{\text{true},k-1}$ is the estimated acceleration, $\tilde{f}_k$ is the measured acceleration and $b_{k-1}$ is the calculated bias. $\text{Error}_{k-1}$ is the error term, as derived from the gravitational constraint. ${b}_{i,k}$ are the bias estimates along each axis $i \in \{x,y,z\}$ from the previous iteration.

The process continues iteratively, updating the estimated biases using these corrections until convergence is reached, defined as when the magnitude of bias corrections falls below a predefined tolerance, or a maximum number of iterations is completed.

\section{Proposed Approach}

Our core motivation is to develop a practical alternative to standard stationary accelerometer calibration techniques. Before presenting the proposed approach, we begin with a motivation and show why an orientation-free calibration approach is needed. 
\subsection{Motivation}
In real-world scenarios, accelerometer calibration is limited to situations where the accelerometers are leveled as even small roll or pitch angles introduce gravity-related offsets that are indistinguishable from true bias. To quantify this effect, an analytical evaluation was conducted to determine the maximum allowable orientation for which the stationary assumption holds and a simple calibration procedure could be applied. 

Based on  (\ref{acc model_angle}) The projected gravity vector components are:
\begin{equation}\label{angle_1}
f(\theta,\phi)_x = \sin(\theta) \cdot g
\end{equation}
\begin{equation}\label{angle_2}
f(\theta,\phi)_y = -\sin(\phi) \cos(\theta) \cdot g
\end{equation}
\begin{equation}\label{angle_3}
f(\theta,\phi)_z = -\cos(\phi) \cos(\theta) \cdot g
\end{equation}

In equations (\ref{angle_1})-(\ref{angle_3}), $\theta$ and $\phi$ represent the angles of interest, and g denotes the gravitational constant. Thus, in stationary conditions, for non-zero roll and pitch angles, a constant offset across all samples exists and its indistinguishable from a true bias. This can lead to inaccurate bias estimation, since the calibration model would wrongly attribute orientation-induced offsets (from (\ref{angle_1})-(\ref{angle_3})) to the estimated bias.

\begin{figure}[H]
    \centering
    \includegraphics[width=0.75\linewidth]{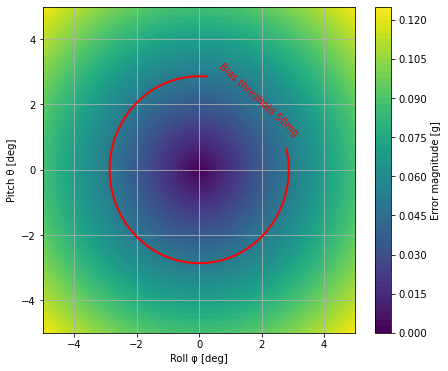}
    \caption{Error magnitude as a function of roll and pitch orientation angles. The red contour indicates the 50mg bias threshold, showing that for combined orientations within approximately ±2.87°, the orientation-induced error remains smaller than a typical accelerometer bias.}.\label{error_smallangles}
\end{figure}

To quantify the validity range, we computed the absolute deviation between the leveled stationary output (zero roll and pitch angles) and the rotated output as a function of $\theta$ and $\phi$. This deviation was treated as an effective bias-like error. The maximum allowable pitch and roll angles for stationary calibration were defined as the point where this error is less than 50mg. Beyond this threshold, the assumption of a leveled accelerometer breaks down, and the simplified model no longer distinguishes between tilt and bias. In such cases, stationary calibration is not available and other complex procedures are required.
As shown in Figure \ref{error_smallangles}, for angles up to approximately 2.87°, the rotation-induced error remains smaller than 50mg. Therefore, small angles can be tolerated, and calibration methods can safely use the bias-dominant model under leveled conditions. Of course, for accurate performance one will require a bias value smaller than 50mg. Thus, the allowable angles will also be smaller, limiting even more real world calibration scenarios.

\subsection{Background}
These methods typically require placing the sensor in multiple distinct orientations so that the known gravity direction can serve as a reference for estimating calibration parameters. However, such procedures often involve an extended calibration duration and careful handling of the sensor’s pose. Other calibration approaches utilize the physical constraint that, under stationary conditions, the specific force magnitude equals the local gravity magnitude. Yet, such model-based approaches suffer from the same drawbacks, as they still require multiple carefully executed orientations and long calibration sessions.

To avoid these shortcomings, we propose an orientation-free neural-based accelerometer bias estimation approach. Specifically, we developed an end-to-end orientation-free bias estimation neural network, OFBENet, that predicts the accelerometer’s bias vector directly from raw stationary time-series readings, without requiring prior knowledge of the sensor’s orientation. Neural networks are well suited to this task, as they combine the ability to reduce noise with the capacity to model nonlinear behavior. Therefore, OFBENet is designed to leverage the stationary nature of the accelerometer signal, where the bias can be assumed constant over short recording periods. This enables the network to capture patterns that persist over time while distinguishing them from sensor noise.

By training a model to learn the mapping of stationary segments of raw accelerometer time-series data to known bias values, the model effectively captures it spatial and temporal signal features that encode the underlying bias even in the presence of measurement noise. Once trained, the network can generalize to unseen data and infer biases in short recordings. This enables fast, scalable, and low-cost calibration for a wide range of MEMS-based accelerometers.

\begin{figure}[H]
    \centering
    \includegraphics[width=1\linewidth]{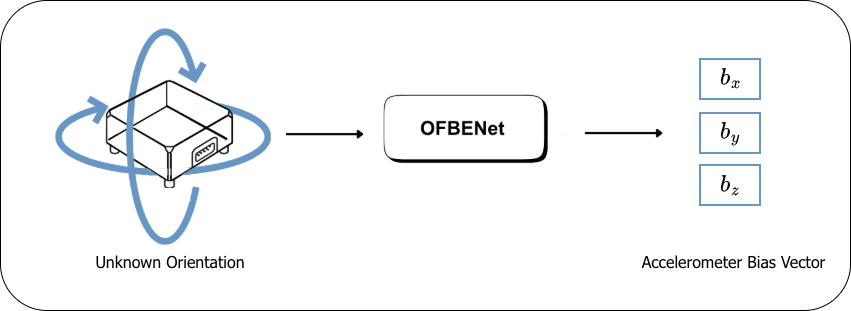}
    \caption{Overview of the proposed calibration method. The pipeline begins with a stationary orientation-free accelerometer readings, which is fed into our simple, yet efficient, neural network OFBENet. The output of the network is the accelerometers deterministic bias vector.}\label{overview}
\end{figure}

An overview of the proposed approach is shown in Figure~\ref{overview}. It illustrates the data flow from raw stationary accelerometer signals (in an unkown rotation) to predicted accelerometer bias vector via our OFBENet approach.

\subsection{Network Architecture}
A lightweight 1D neural network was designed to prioritize generalization and efficiency. The input, accelerometer readings, are windowed into fixed-length segments. A non-overlapping windowing strategy (stride equal to window size) was used to ensure that each segment remains independent and non-redundant. 

The input to the network is a sequence of accelerometer measurements:
\begin{equation}
\mathbf{X} \in \mathbb{R}^{T \times 3},
\label{eq:cnn_input}
\end{equation}
where $T$ is the number of samples and each row contains the 3D components of the specific force vector. Therefore, each input segment has a shape of $(T,3)$, corresponding to the three raw accelerometer axes.

\begin{figure}[H]
    \centering
    \includegraphics[width=1\linewidth]{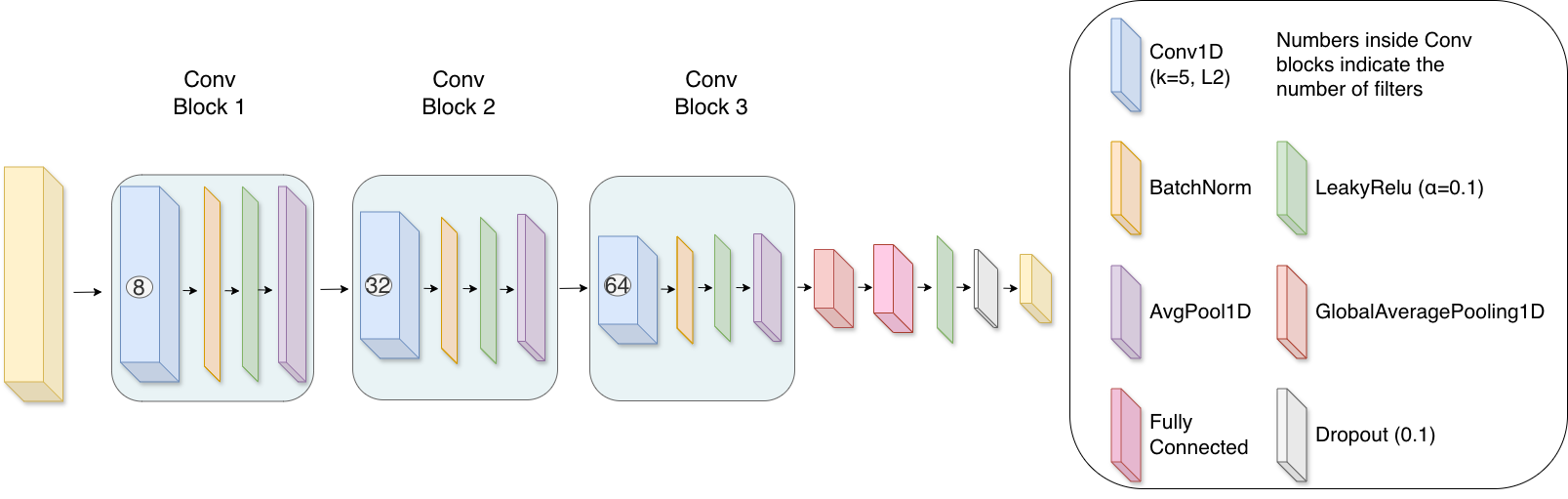}
    \caption{OFBENet architecture used for bias estimation. Each convolutional block includes a 1D convolutional layer, batch normalization, Leaky ReLU activation, and average pooling. The network concludes with global average pooling, a fully connected layer, dropout, and an output layer.}.
    \label{cnn_diagram}
\end{figure}

The architecture includes three convolutional layers with kernel size $m=5$, capturing increasingly complex temporal patterns using filter depths of 8, 32, and 64, respectively. Each convolution is followed by batch normalization and a LeakyReLU activation, forming a feature extraction block. The OFBENet architecture is shown in Figure ~\ref{cnn_diagram}.

The 1D convolution operation is defined as:
\begin{equation}
Z_{i,k} = \sum_{j=0}^{m-1} \sum_{c=1}^{3} x_{i+j,\,c}\, w_{j,c,k} + b_k
\label{eq:cnn_1d_conv}
\end{equation}
where $Z_{i,k}$ denotes the output at time index $i$ and channel $k$, $w_{j,c,k}$ and $b_k$ are the convolution parameters.
The convolution extracts local temporal features from the accelerometer signals, but the resulting activations may vary in scale across filters and training batches. 
To maintain consistent feature magnitudes and stabilize learning, batch normalization was applied after each convolutional layer \cite{bjorck2018understandingbatchnormalization}: 

\begin{equation}
\mathrm{BN}(Z_{i,k}) = \gamma_k \frac{Z_{i,k} - \mu_k}{\sqrt{\sigma_k^2 + \epsilon}} + \beta_k
\label{eq:batchnorm}
\end{equation}
where $\mu_k$ and $\sigma_k^2$ are the batch mean and variance for channel $k$, respectively, $\gamma_k$ is a learnable scale parameter, $\beta_k$ is a learnable shift parameter, and $\epsilon$ is a small constant for numerical stability.

To introduce nonlinearity, the LeakyReLU activation function \cite{maas2013rectifier} was used in all convolutional layers. It is defined as:
\begin{equation}
\sigma_{\text{LeakyReLU}}(Z_i) =
\begin{cases}
Z_i, & Z_i \ge 0, \\[6pt]
\alpha Z_i, & Z_i < 0
\end{cases}
\label{eq:leakyrelu}
\end{equation}
where $\alpha = 0.1$. Leaky ReLU is an improved variant of the standard ReLU function that addresses the "dying ReLU" problem by introducing a small linear component for negative input values, rather than setting them to zero \cite{sharma2017activation}. This activation function
maintains gradient flow during backpropagation by providing a small but non-zero gradient for negative inputs, preventing neurons from becoming permanently inactive when weights and biases would otherwise not be updated. The choice of $\alpha = 0.1$ was found beneficial for convergence in preliminary experiments, avoiding the gradient saturation issues commonly encountered with standard ReLU in deep networks.

Following activation, each block applies average pooling to downsample the temporal dimension using:
\begin{equation}
Y_{i,k} = \frac{1}{P}\sum_{r=0}^{P-1} Z_{i\cdot s_p + r,\,k}
\label{eq:avgpool}
\end{equation}
where $P$ is the pooling window size. Average pooling provides a generalized representation of the signal by smoothing local variations and reducing noise \cite{nirthika2022pooling}.

After the three convolutional blocks, global average pooling reduces each feature map to a single scalar:
\begin{equation}
h_k = \frac{1}{L_k} \sum_{i=1}^{L_k} Y_{i,k}
\label{eq:gap}
\end{equation}
where $L_k$ is the temporal length of feature map $k$. In this network, the final convolutional layer produces 64 feature maps, resulting in 64 averaged values after global pooling. The output $\mathbf{h} = [h_1, h_2, \ldots, h_{64}]$ forms a compact global representation of the input window, summarizing the overall activation of each feature learned over time.

This vector serves as the input to the dense layer, which combines the globally pooled features to estimate the final bias components. The dense layer performs a linear transformation on the global feature vector:
\begin{equation}
z_{j} = \sum_{k=1}^{n} h_{k}\, w_{k,j} + b_{j}
\label{eq:dense}
\end{equation}
where $\mathbf{h} \in \mathbb{R}^n$ is the input from the global average pooling layer, and $w_{i,j}$ and $b_j$ are the trainable weights and biases of neuron $j$.

Dropout is applied using the following modified formulation \cite{Goodfellow-et-al-2016}: 
\begin{equation}
{h}_{drop} = \frac{d_i}{p} \, h_i , \qquad d_i \sim \mathrm{Bernoulli}(p)
\label{eq:dropout}
\end{equation}
where $p$ is the inclusion probability. This ensures that the correct expected value is captured. 

Finally, a linear output layer produces the predicted bias vector:
\begin{equation}
\hat{{y}} = {W}_{\text{out}} {h}_{\text{drop}} + {b}_{\text{out}}
\label{eq:output}
\end{equation}
where ${h}_{\text{drop}}$ is the dropout-regularized dense output, and ${W}_{\text{out}}$, ${b}_{\text{out}}$ are trainable parameters. The final output $\hat{{y}} = [\hat{b}_x, \hat{b}_y, \hat{b}_z]$ represents the predicted bias vector along the three accelerometer axes.

\subsection{Training Process}
OFBENet was trained using the mean squared error (MSE) loss function, which is appropriate for continuous-valued regression tasks such as bias estimation. The MSE is defined as:

\begin{equation}
\text{MSE} = \frac{1}{n} \sum_{i=1}^{n} (b_i - \hat{b}_i)^2
\end{equation}

where $n$ is the number of samples, $b_i$ is the true bias vector, and $\hat{b}_i$ is the predicted bias vector output by OFBENet.

To optimize this loss, the Adam optimizer was employed with a learning rate of 0.01. Adam was selected as the optimizer due to its ability to adapt the learning rate for each parameter individually and handle gradient noise effectively \cite{kingma2014adam}. A batch size of 8 was used to improve generalization and reduce overfitting, as larger batch sizes were found to yield less stable results in early experiments.

A learning rate adaptation strategy was employed, in which the learning rate was automatically reduced  by a factor of 0.2 when the validation loss failed to improve for 15 consecutive epochs.  This mechanism helps the optimizer escape shallow minima when convergence stagnates. Additionally, early stopping with a 5-epoch patience threshold was applied following established practices for overfitting prevention \cite{Prechelt1998}. To determine OFBENet parameters during training, OFBENet was trained on a commercial laptop, MacBook Pro (16-inch, 2019). It has a 2.6 GHz 6-core Intel Core i7 CPU, 32 GB of DDR4 RAM, and an integrated Intel UHD Graphics 630 GPU with 1536 MB VRAM. The code implementation of OFBENet was done using Python’s “TensorFlow 2.16.2” library.

\section{Analysis and Results}
We begin by describing the datasets recorded for this research, followed by analysis and results of our proposed approach.

\subsection{Datasets}
Three datasets were used in this research to evaluate and compare the performance of model-based and learning-driven calibration methods for stationary accelerometers.

The first dataset consists of accelerometer recordings from four SparkFun IMUs positioned in a gravity-aligned (zero roll and pitch angles) orientation. This dataset is used to analyze the bias convergence behavior, determine the minimum number of samples required for reliable bias estimation, and evaluate OFBENet’s prediction accuracy.

The second dataset includes accelerometer recordings from two SparkFun IMUs placed in different orientations. This dataset allows offers a more real world practical scenarios. 

\subsubsection{Gravity-Aligned Dataset}
\begin{figure}[H]
    \centering
    \includegraphics[width=0.5\linewidth]{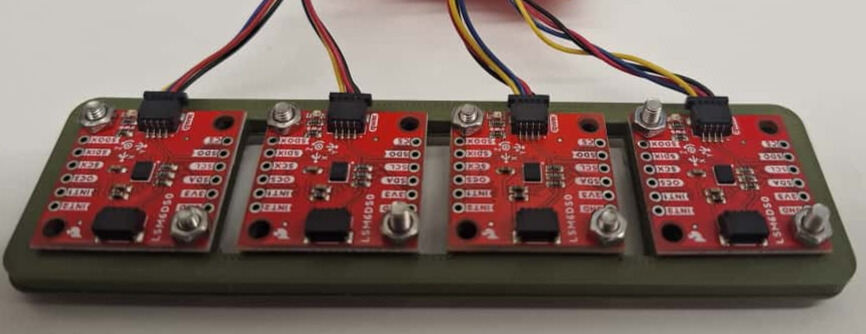}
    \caption{Four SparkFun ADXL345 accelerometers used in the gravity-aligned dataset, placed on a leveled surface.}
    \label{fig:placeholder}
\end{figure}

This dataset was collected using four SparkFun ADXL345 IMUs \cite{sparkfun2024} placed in a stationary, gravity-aligned orientation as shown in Figure \ref{fig:placeholder}. Under these conditions, the accelerometer’s sensitive axes are aligned with the global reference frame and the ideal accelerometer output is the local gravity magnitude in a single axis, while the other two axes should have zero readings. Before data collection, the sensors' orientation was verified to ensure that both roll and pitch angles were approximately zero. The SparkFun IMUs were placed on a stable, level surface.

The main objective of this dataset was to determine the minimum number of samples required to reliably estimate the bias by simple averaging, as well as to evaluate OFBENet under low-variation conditions. This calculation was based on data collected from all four accelerometers.
This dataset also served as a clean benchmark for testing OFBENet in a  bias-dominated regime. For a single IMU, 100 recordings were made, each lasting 80 seconds, resulting in a total of 400 recordings with a total duration of 8.89 hours.

\subsubsection{Rotated Accelerometer Dataset}
This dataset was collected using two SparkFun ADXL345 IMUs alongside a high-grade Memsense MS-3025 IMU \cite{memsense_msimu3025} that served as the ground truth (GT) reference. 

\begin{figure}[H]
    \centering
    \begin{subfigure}[t]{0.45\linewidth}
        \includegraphics[width=\linewidth]{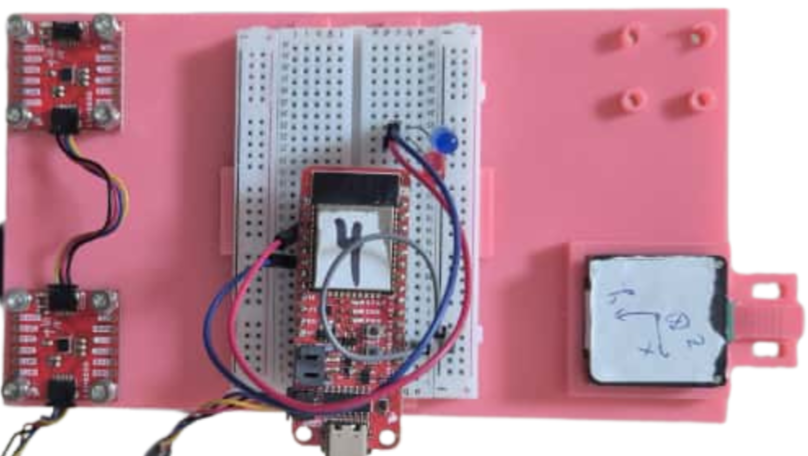}
        \caption{} % subcaption for (a)
        \label{fig:top_view}
    \end{subfigure}
    \hfill
    \begin{subfigure}[t]{0.25\linewidth}
        \includegraphics[width=\linewidth]{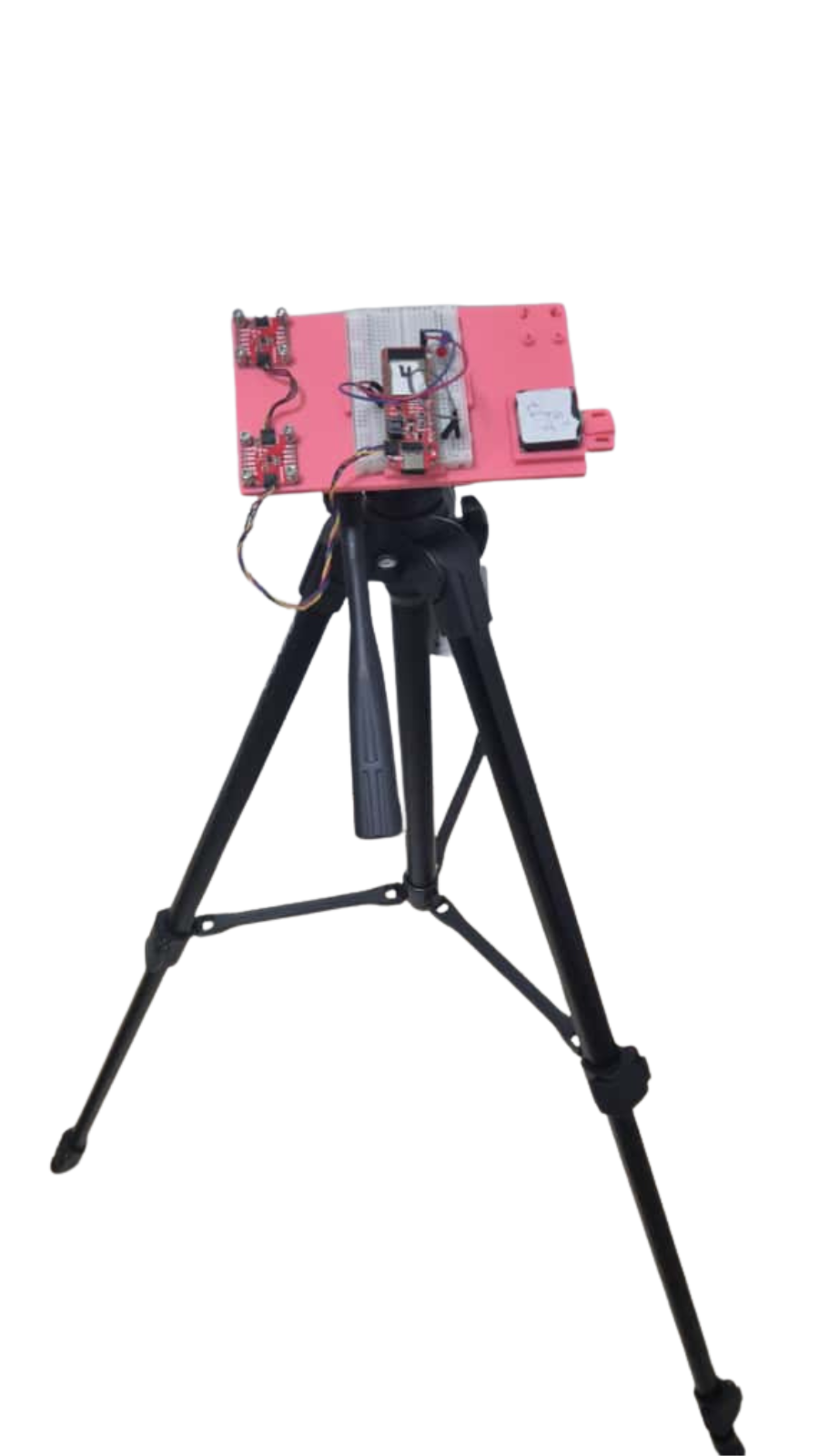}
        \caption{} % subcaption for (b)
        \label{fig:tripod}
    \end{subfigure}
    \caption{Experimental setup used for recording the rotated accelerometer dataset. 
    (\textbf{a}) Top view of the 3D-printed plate showing the fixed alignment of the IMUs. 
    The sensor on the right is the Memsense MS-3025, and the two sensors on the left are SparkFun ADXL345 IMUs connected to the SparkFun development board. 
    (\textbf{b}) Experimental setup mounted on a tripod used for orientation adjustment during data collection.}
    \label{fig:setup_combined}
\end{figure}

All three IMUs were mounted on a custom 3D-printed plate designed to hold the sensors in a fixed and stable alignment as shown in Figure \ref{fig:top_view}.
The plate was attached to a tripod in place of the standard camera mount, allowing controlled changes in pitch and roll angles by adjusting the tripod’s tilt levers. This setup is presented in Figure \ref{fig:tripod}.
The SparkFun IMUs were operated via a SparkFun Thing Plus ESP32 WROOM development board \cite{sparkfun2024esp32} running custom Arduino firmware, while the Memsense IMU was connected to its dedicated software suite \cite{memsense2024mscip}. 
The two SparkFun sensors were positioned side-by-side below the Memsense unit to ensure parallel sensing axes.

\begin{figure}[H]
    \centering
    \includegraphics[width=0.95\linewidth]{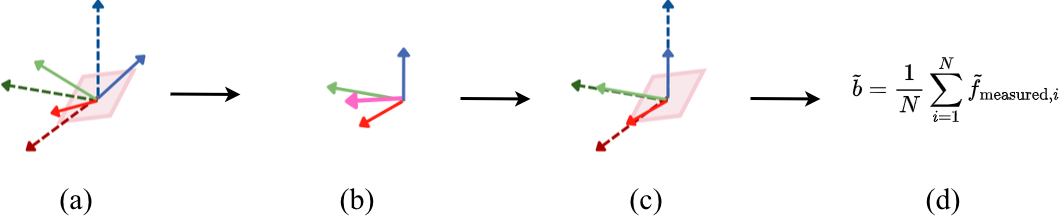}
    \caption{Bias extraction pipeline used to label the rotated accelerometer dataset. The Memsense IMU provides reference orientation for each recording, allowing accurate bias estimation of the SparkFun accelerometer signals through alignment and mean computation. (\textbf{a}) Simultaneous data recording on the 3D-printed plate with SparkFun and Memsense IMUs at multiple pitch and roll angles. (\textbf{b}) Orientation vector extraction from the Memsense IMU. (\textbf{c}) Alignment of SparkFun IMUs by applying the inverse rotation matrix. (\textbf{d}) Mean computation to estimate the accelerometer bias, used as the ground-truth label.}

    \label{fig:bias_extraction}
\end{figure}

The Memsense IMU, with a superior bias stability of ±0.395~mg compared to the SparkFun’s ±20~mg, was used as the reference sensor for determining the true orientation of each recording. 
During post-processing, the Memsense accelerometer readings were used to compute the rotation matrix corresponding to each orientation using Rodrigues’ rotation formula \cite{corke2023robotics}. 
This rotation matrix was then applied to the SparkFun recordings to align them with the global reference frame. 
After alignment, the bias of each SparkFun sensor was computed as the mean of the aligned signal, as illustrated in Figure~\ref{fig:bias_extraction}. This calculation was made at the end of the recording long after the bias value converged. 

Before collecting the full dataset, a validation test confirmed that the bias remained stable over time: the sensors were recorded at nearly zero roll and pitch angles, tilted through three different orientations, and returned to zero without powering off. The first and last recordings were then compared, confirming that the bias remained consistent throughout the session. 

During dataset collection, the plate was rotated to different orientations covering approximately $-80^\circ$ to $+60^\circ$ in pitch and $-180^\circ$ to $+180^\circ$ in roll. The dataset includes recordings with isolated pitch or roll rotations, as well as combined pitch–roll orientations, ensuring that the calibration methods were evaluated under both single-axis and multi-axis conditions.
For every three consecutive recordings, the IMUs were powered off and restarted to generate a new turn-on bias. As a result, each bias instance corresponds to three unique angular orientations. 
The complete dataset includes 87 recordings per SparkFun sensor, each captured in a different orientation (2.25 hours total per device), providing diverse orientation data with controlled bias variability. Each recording comprised 12,000 samples (80 seconds), whereas for the test signals, only 3,000 samples (20 seconds) were required for reliable evaluation and were therefore used in the analysis. The two accelerometers were used independently across the calibration methods, rather than combined into a single dataset, to verify that OFBENet generalizes effectively across different devices.

It is important to note that accurate bias estimation requires precise alignment between the SparkFun and Memsense sensors. 
As shown in the small-angle analysis (Section~\ref{stationary_readings}), misalignments larger than approximately 2.87° can introduce gravity-induced deviations that are indistinguishable from true bias. 
The 3D-printed plate minimized these errors by maintaining the sensors within this threshold throughout all recordings.

\subsubsection{Summary}
\label{summary}
Table \ref{tab:dataset_times} summarizes the collection of datasets, including accelerometer signals from multiple IMU sensors with a combined recording duration of 13.39 hours. It also shows our division to training/test and the number of recordings in each set. Each recording lasted between 80 and 93 seconds. The number of samples in each dataset is calculated by:

\begin{equation}
N = t_{\text{hours}} \times 3600 \times f_s
\end{equation}

where $t_{\text{hours}}$ is the recording duration in hours and $f_s = 150\,\text{Hz}$ is the sampling rate. In total, six different SparkFun IMUs were used in the dataset.

\begin{table}[H]\centering
\caption{Recording durations and counts per dataset.}
\label{tab:dataset_times}
\resizebox{\linewidth}{!}{%
\begin{tabular}{|l|c|c|c|c|c|c|c|}
\hline
\multirow{2}{*}{\textbf{Dataset}} & \multirow{2}{*}{\textbf{\# Sensors}} & \multicolumn{2}{c|}{\textbf{Train}} & \multicolumn{2}{c|}{\textbf{Test}} & \multicolumn{2}{c|}{\textbf{Total}} \\
\cline{3-8}
& & \textbf{Time [hr]} & \textbf{\#Rec} & \textbf{Time [hr]} & \textbf{\#Rec} & \textbf{Time [hr]} & \textbf{\#Rec} \\
\hline
Gravity-Aligned Dataset & 4 & 6.22 & 280 & 2.67 & 120 & 8.89 & 400 \\
\hline
\shortstack[l]{Rotated Accelerometer\\Dataset} & 2 & 4.04 & 156 & 0.46 & 18 & 4.50 & 174 \\
\hline
Total & 6 & \textbf{10.26} & \textbf{436} & \textbf{3.13} & \textbf{138} & \textbf{13.39} & \textbf{574} \\
\hline
\end{tabular}}
\end{table}

\subsection{Evaluation Metrics}
We employed three evaluation metrics: 1) root mean square error (RMSE) 2) maximum error, and 3) $t$-test.

The RMSE was chosen to enable direct comparison across the different methods (baseline and ours). The RMSE is defined as:

\begin{equation}
\mathrm{RMSE} = \sqrt{\frac{1}{N} \sum_{i=1}^{N} \left\| \mathbf{y}_i - \hat{\mathbf{y}}_i \right\|^2}
\label{eq:rmse}
\end{equation}

\noindent
where $N$ is the number of test signals, $y_i$ represents the true bias vector from each test signal, and $\hat{y}_i$ is the estimated bias vector. RMSE retains the same physical units as the bias, making the results directly interpretable.

To capture the worst-case deviation from the GT, the maximum error was also calculated as:

\begin{equation}
\label{eq:max}
\mathrm{E_{max}} = \max_{1 \leq i \leq N} |\mathbf{y}_i - \mathbf{\hat{y}}_i|
\end{equation}

\noindent
In addition, a paired two-sided $t$-test was conducted to evaluate the statistical significance of performance differences between methods. The $t$-statistic is computed as:

\begin{equation}
t = \frac{\bar{d}}{s_d / \sqrt{N}}
\label{eq:ttest}
\end{equation}

\noindent
where $\bar{d}$ is the mean of the error differences between two methods and $s_d$ is their standard deviation. 
The resulting $p$-value, derived from the $t$-distribution with $N-1$ degrees of freedom, represents the probability of observing a difference as large as (or larger than) the measured one under the null hypothesis that both methods perform equally. 
A $p$-value below 0.05 is typically considered statistically significant, indicating that the observed difference is unlikely to have occurred by chance.

\subsection{Bias Convergence Analysis}
\label{subsubsec:bias_estimation}

To determine the minimum number of samples required for reliable bias estimation, a convergence analysis was conducted on the gravity-aligned accelerometer dataset. To avoid device-specific effects, the convergence analysis used all recordings from all four SparkFun sensors. For each signal, the running mean was cumulatively calculated over time, and convergence was defined as the point where the derivative of the last 40 running means dropped below $5\times10^{-6}\,\mathrm{m/s^2}$, indicating that the signal’s average had converged.

\begin{figure}[h!]
    \centering
    \begin{subfigure}[b]{0.48\linewidth}
        \includegraphics[width=\linewidth]{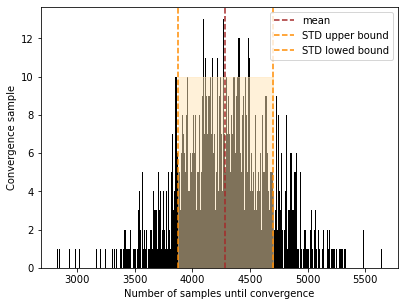}
        \caption{} % subcaption for (a)
        \label{fig:convergence_hist}
    \end{subfigure}
    \hfill
    \begin{subfigure}[b]{0.48\linewidth}
        \includegraphics[width=\linewidth]{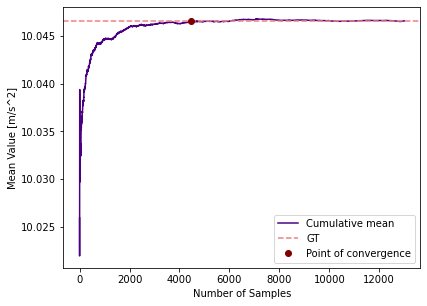}
        \caption{} % subcaption for (b)
        \label{fig:convergence_example}
    \end{subfigure}
    \caption{Convergence analysis of the stationary accelerometer dataset used to determine the minimum number of samples required for reliable bias estimation. 
    (\textbf{a}) Histogram of samples required for convergence across all signals. 
    (\textbf{b}) Example of a convergence curve for a single signal, showing the reference ground-truth bias.}
    \label{fig:convergence_analysis}
\end{figure}

The histogram in Figure ~\ref{fig:convergence_hist} summarizes the number of samples required to reach convergence across all .400 signals from the gravity-aligned dataset, evaluated along the X axis. The X axis is shown as a representative case, since the Y and Z axes demonstrated similar convergence behavior. The mean number of samples until convergence was approximately 4500, with most signals converging between 4000 and 5000 samples. This range is visualized by the shaded region, bounded by one standard deviation from the mean. An example of a convergence curve for a single recording is shown in Figure~ \ref{fig:convergence_example}, illustrating how the running mean stabilizes over time for a single stationary signal. As a result of this analysis, the bias labels for all datasets were defined as the mean of the signal after 4500 samples.

\subsubsection{Bias Estimation Across Datasets}
To ensure a statistically robust evaluation, all calibration methods were evaluated using a five-fold cross-validation procedure adapted to the structure of each dataset.

In the gravity-aligned dataset, the recordings were randomly divided into training and testing subsets, with 20\% of the data (80 signals) used for testing in each fold. The test recordings were randomly drawn from all sensors, which ensured that each fold included a representative and balanced subset of measurements from each device. Across all datasets, the test sample length was 3000 samples (20 seconds). 

In the rotated accelerometer datasets, each fold used a different combination of nine test signals, evaluated five separate times. In this case, the data was split across different tilt angles, ensuring that every model was tested on unseen orientations rather than repeated conditions.

Table \ref{tab:rmse_results} presents the cross-validation results for all datasets, comparing the two model-based baselines with our approach. As shown by the mean and standard deviation (STD) of the RMSE across folds, OFBENet consistently achieved the lowest mean RMSE across all datasets. In the rotated dataset, OFBENet achieved an improvement of more than 77\% compared to the model-based approaches. The slightly higher RMSE observed in the gravity-aligned dataset can be attributed to the limited variability of the input signals and the relative scaling of the target values. In this configuration, the accelerometer remained stationary and level with respect to gravity, resulting in measurements that exhibit minimal temporal and spatial variation. Consequently, the network was exposed to a weaker learnable signal compared to the tilted datasets, in which orientation changes introduce richer input dynamics. Under such low-variability conditions, the convolutional layers have limited structural information to extract, which may slightly constrain the network’s ability to capture fine-grained bias patterns.
\begin{table}[H]

\centering
\caption{Cross-validation dataset comparison between the model-based baselines and our approach.}
\label{tab:rmse_results}
\begin{adjustbox}{max width=\textwidth}
\renewcommand{\arraystretch}{1.15}
\begin{tabular}{|l|c|l|c|c|c|}
\hline
\textbf{Dataset} & \textbf{\# Sensors} & \textbf{Method} & \textbf{Mean RMSE (m/s\textsuperscript{2})} & \textbf{STD (m/s\textsuperscript{2})} & \textbf{Improv (\%)} \\
\hline
\multirow{2}{*}{Gravity-Aligned Dataset} 
    & \multirow{2}{*}{4} & Least Squares       & 0.1767 & 0.0048 & 52.52 \\
    &                    & \textbf{OFBENet (ours)} & 0.0839 & 0.0197 & -- \\
\hline
\multirow{3}{*}{\shortstack[l]{Rotated Accelerometer\\Dataset - SparkFun 1}} 
    & \multirow{3}{*}{1} & Least Squares       & 0.1711 & 0.1039 & 77.11  \\
    &                    & Mathematical Model  & 0.2088 & 0.1243 & 81.24  \\
    &                    & \textbf{OFBENet (ours)} & 0.0392 & 0.0462 & -- \\
\hline
\multirow{3}{*}{\shortstack[l]{Rotated Accelerometer\\Dataset - SparkFun 2}} 
    & \multirow{3}{*}{1} & Least Squares       & 0.3180 & 0.2506 & 87.19  \\
    &                    & Mathematical Model  & 0.1869 & 0.0583 & 78.21  \\
    &                    & \textbf{OFBENet (ours)} & 0.0407 & 0.0189 & -- \\
\hline
\end{tabular}
\end{adjustbox}
\label{tab:cv_rmse_results}
\end{table}

The small differences between the two tilted SparkFun accelerometers can be attributed to slight mounting imperfections. Although the 3D-printed plate ensured a fixed configuration, minor angular deviations of about 1.5° between sensors and relative to the Memsense reference remained. As discussed in the small-angle analysis in Section \ref{stationary_readings}, deviations below 2.87° are indistinguishable from true bias and, therefore, do not compromise calibration accuracy. The observed discrepancies are within acceptable bounds and confirm the robustness and consistency of the proposed approach.

\begin{figure}[h!]
    \centering
    \includegraphics[width=1\linewidth]{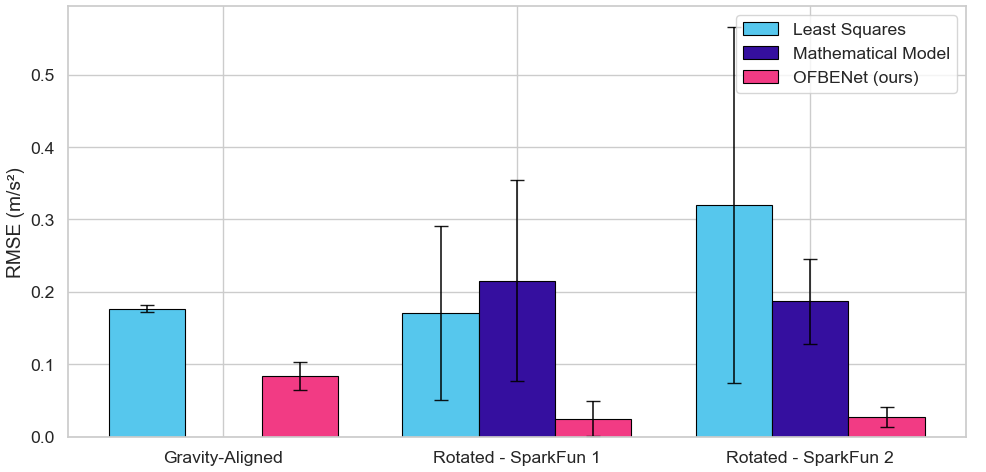}
    \caption{Mean RMSE and STD (of the five fold cross validation) for each calibration approach and across all datasets.}
    \label{fig:rmsemean}
\end{figure}

Figure \ref{fig:rmsemean} visualizes the mean RMSE values and standard deviations obtained by five-fold cross-validation. OFBENet shows both the lowest average RMSE and the smallest variance, indicating stable and reliable generalization. In contrast, the least-squares and mathematical-model methods exhibit greater variability, reflecting their dependence on orientation diversity and data partitioning. Although all methods experience slightly higher errors in the rotated datasets, OFBENet’s degradation is minimal, confirming its robustness to tilt-induced variability.

\begin{figure}[h!]
    \centering
    \includegraphics[width=1\linewidth]{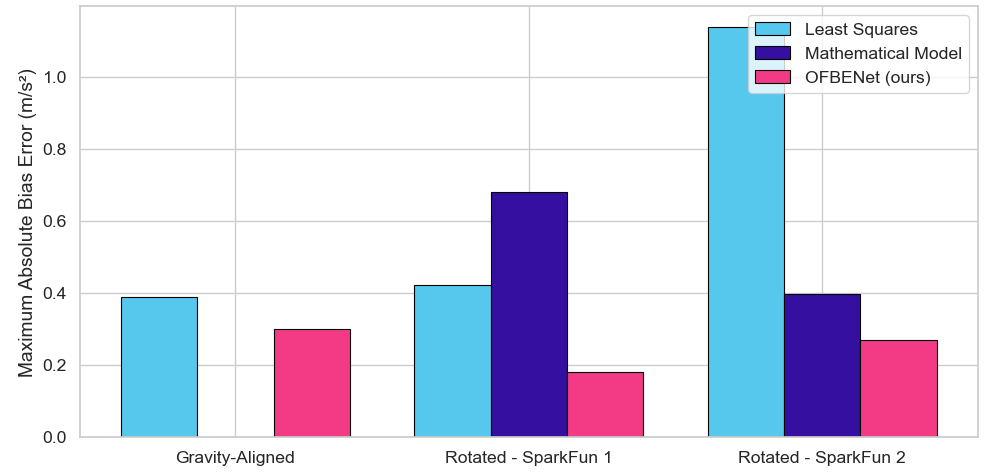}
    \caption{Maximum absolute bias error for each calibration approach and across all datasets.}
    \label{fig:max error}
\end{figure}

Figure \ref{fig:max error} shows the maximum absolute bias errors obtained for each calibration method and dataset using (\ref{eq:max}). OFBENet consistently yielded the smallest maximum errors, indicating a more stable and bounded estimation performance. In the gravity-aligned dataset, the differences between methods were modest due to the limited signal dynamics, yet OFBENet still achieved slightly lower extremes. In the rotated datasets, the least-squares approach exhibited the largest errors, reflecting its sensitivity to orientation and noise, while the mathematical model showed moderate improvements. In contrast, OFBENet maintained low maximum errors across both SparkFun sensors, with an improvement of more than 32.38\%, further demonstrating its robustness to tilt variations and its ability to suppress large bias outliers.

\begin{figure}[h!]
    \centering
    \includegraphics[width=1\linewidth]{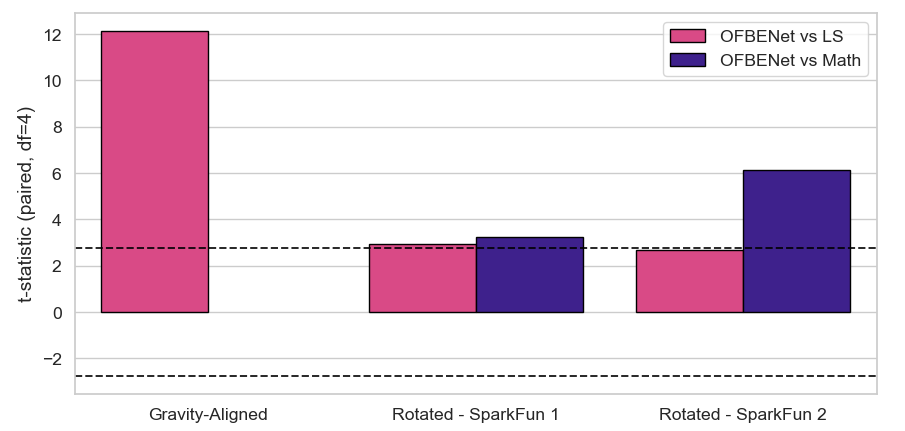}
    \caption{paired t-test results comparing OFBENet with the baseline calibration methods. The dashed lines denote the critical t-values for a two-tailed test with four degrees of freedom ($\alpha=0.05$). }
    \label{fig:ttest}
\end{figure}

Figure \ref{fig:ttest} shows the results of the paired t-test. In the gravity-aligned dataset, the improvement over the least-squares approach is highly significant. In both rotated datasets, the comparisons with the least-squares and mathematical model methods also produce t-values that exceed the critical threshold, confirming that OFBENet’s lower RMSE is statistically significant. These results validate that the observed improvements are consistent across folds and not due to random variability.

\section{Conclusions}
This paper presented a neural network approach to calibrating the bias of low-cost stationary accelerometers. The accelerometers used in this paper are representative of consumer-grade sensors, which typically exhibit greater bias instability due to manufacturing variability and environmental sensitivity. These factors make accurate and robust calibration particularly challenging, especially in practical settings where external equipment or multi-orientation sequences are unavailable.

OFBENet achieved high accuracy and demonstrated strong generalization. In particular, it performed consistently across two distinct accelerometer readings from SparkFun IMUs, indicating that it does not simply memorize device-specific patterns, but rather learns a generalizable representation of accelerometer bias. This robustness across hardware instances highlights the practicality of the approach for deployment in real-world conditions.

Comprehensive five-fold cross-validation confirmed that OFBENet achieved the lowest RMSE across all datasets, with reduced variance between folds. Paired t-tests further showed that these improvements were statistically significant ($p<0.05$) relative to both least-squares and mathematical model-based methods. In addition, OFBENet maintained the smallest maximum bias errors, demonstrating stability and bounded estimation performance even under tilted configurations. Together, these results establish the statistical reliability and robustness of the proposed approach.

The statistical results in the gravity-aligned dataset highlight a key distinction: while the least-squares method depends on orientation diversity to achieve stable calibration, OFBENet maintained low and consistent errors even in the absence of such variation. This demonstrates that the network captures intrinsic bias patterns directly from the signal distribution rather than relying on multi-position information. OFBENet therefore removes the need for multi-position calibration, simplifying practical implementation.

Finally, while iterative and least-squares methods rely on carefully designed mathematical formulations, they are also highly sensitive to numerical conditioning, rank deficiencies, and orientation diversity. In contrast, OFBENet does not depend on such matrix inversions or iterative parameter updates. Once trained, it directly maps raw accelerometer signals to bias estimates, bypassing the numerical instabilities inherent in traditional methods. This not only explains the lower error metrics observed, but also underscores the robustness of data-driven approaches in scenarios where analytical models become ill-conditioned. These findings are consistent with previous work on neural network–based gyroscope calibration, further supporting the use of data-driven methods for rapid and accurate calibration in low-cost accelerometers.

The high statistical accuracy and consistency demonstrated by OFBENet highlight its potential for precision-demanding applications such as navigation, robotics, and autonomous systems, where precise bias calibration is critical for maintaining long-term stability and minimizing drift. By providing accurate and stable bias estimation in an unknown orientation, the proposed method effectively overcomes one of the main limitations of these sensors. This improvement enables their reliable integration into systems, reducing cost without compromising accuracy. 

\section*{Acknowledgment}

M. L. is supported by the Maurice Hatter foundation scholarship.
\newpage
% ------------- Must have for the references -------------------
\renewcommand\refname{References} % name for the reference list
{\setstretch{1.0} % linespacing for the references
\addcontentsline{toc}{section}{References} % to change the name of the references in the TOC
\bibliography{References} % adds the references to the document
\bibliographystyle{IEEEtranN}
}
\end{document}